\documentclass[conference]{IEEEtran}
\usepackage[utf8]{inputenc}
\IEEEoverridecommandlockouts
\usepackage{cite}
\usepackage{float}
\usepackage{xcolor}

\usepackage{footmisc}
\usepackage{listings}

\usepackage{multicol}
\usepackage{marvosym} 
\usepackage{adjustbox}
\usepackage[utf8]{inputenc}

\usepackage{url}
\usepackage[section]{placeins}
\usepackage{amsmath,amssymb,amsfonts}
\usepackage{multirow}
\usepackage{algorithm}
\usepackage{algpseudocode}
\usepackage{multicol}
\usepackage{balance}
\usepackage{amsmath}
\usepackage{booktabs}
\usepackage{caption}
\usepackage{graphicx}
\usepackage{hyperref}
\hypersetup{
    colorlinks=true,      
    linkcolor=red,       
    anchorcolor=black,    
    citecolor=blue,      
    filecolor=cyan,       
    menucolor=red,        
    runcolor=cyan,        
    urlcolor=magenta      
}
\usepackage{subcaption}
\usepackage{textcomp}
\usepackage{xcolor}
\usepackage{fancyhdr}
\usepackage{tikz}
\usepackage{pifont} 

\def\BibTeX{{\rm B\kern-.05em{\sc i\kern-.025em b}\kern-.08em
    T\kern-.1667em\lower.7ex\hbox{E}\kern-.125emX}}

\begin{document}


\title{Dynamic Malware Classification of Windows PE Files using CNNs and Greyscale Images Derived from Runtime API Call Argument Conversion

}

\author{\IEEEauthorblockN{Md Shahnawaz}
\IEEEauthorblockA{\textit{Department of CSE} \\
\textit{NIT Rourkela,} \\
Odisha, India\\
\href{https://orcid.org/0009-0003-2936-246X}{0009-0003-2936-246X}\\
}
\and
\IEEEauthorblockN{Bishwajit Prasad Gond}
\IEEEauthorblockA{\textit{Department of CSE} \\
\textit{NIT Rourkela,}\\
Odisha, India \\
\href{https://orcid.org/0000-0003-3640-0463}{0000-0003-3640-0463}\\
}

\and
\IEEEauthorblockN{Durga Prasad Mohapatra}
\IEEEauthorblockA{\textit{Department of CSE} \\
\textit{NIT Rourkela,}\\
Odisha, India \\
\href{https://orcid.org/0000-0002-4824-7091}{0000-0002-4824-7091}\\
}
}

\maketitle

\begin{abstract}
Malware detection and classification remains a topic of concern for cybersecurity, since it is becoming common for attackers to use advanced obfuscation on their malware to stay undetected. Conventional static analysis is not effective against polymorphic and metamorphic malware as these change their appearance without modifying their behavior, thus defying the analysis by code structure alone. This makes it important to use dynamic detection that monitors malware behavior at runtime. In this paper, we present a dynamic malware categorization framework that extracts API argument calls at the runtime execution of Windows Portable Executable (PE) files. Extracting and encoding the dynamic features of API names, argument return values, and other relative features, we convert raw behavioral data to temporal patterns. To enhance feature portrayal, the generated patterns are subsequently converted into grayscale pictures using a magma colormap. These improved photos are used to teach a Convolutional Neural Network (CNN) model discriminative features, which allows for reliable and accurate malware classification. Results from experiments indicate that our method, with an average accuracy of 98.36\% is effective in classifying different classes of malware and benign by integrating dynamic analysis and deep learning. It not only achieves high classification accuracy but also demonstrates significant resilience against typical evasion strategies.
\end{abstract}

\begin{IEEEkeywords}
API call, Dynamic Analysis, Magma Colormap, Malware Classification, CNN, PE Files, API Call Arguments.
\end{IEEEkeywords}

\section{INTRODUCTION}\label{sec:intro}

As there is an intensifying digitalization of services and infrastructure, the danger of cybercrime is rising in amplitude and sophistication. Organizations are facing an upswing of advanced cyberattacks fueled by geopolitics, vulnerabilities in the supply chain, and new technologies, according to the World Economic Forum's \textit{Global Cybersecurity Outlook 2025}. Industry reports stress there is an imperative for intelligent, responsive security mechanisms due to changing threat vectors \cite{WEF2025}.

Malware detection has become increasingly harder in recent years with advancements in evasion and code obfuscation at an accelerated rate. Polymorphic and metamorphic malware examples modify their code at run time so that traditional static code inspection, as well as signature-based detection, is evaded by them \cite{Hebish2024}. To bypass these shortcomings, dynamic analysis methods have become popular. By observing run-time activities, particularly API call sequences and their arguments at run-time upon running Windows Portable Executable (PE) files, analysts can identify malicious intent with better reliability \cite{Javed2024}.

One of the latest developments in malware analysis is to convert such sequences of API calls to grayscale or RGB images to produce graphical representations that retain temporal as well as behavioral patterns. These images can further be classified by deep learning models, especially by Convolutional Neural Networks (CNNs), which are exceptionally efficient in extracting spatial features from images \cite{Sasikala2024, Peng2024}. CNNs have manifestly excelled in evasive threat detection based on their capability to learn hierarchically from raw inputs.

Along with standard CNN models, certain works explored advanced architectures and hybrid methods. For instance, Musaev et al. blended an optimized CNN with fine-tuning MobileNetV2 to achieve maximum performance on the Malimg dataset, highlighting that methods based on ensembling perform well \cite{Musaev2025}. They proposed MVC-RSN, a CNN approach that aims to improve malware classification under adversarial environments \cite{Wu2024}. A lightweight CNN architecture designed for effective malware detection in resource-constrained IoT environments was also presented by Yuan et al. \cite{Yuan2022}.

Further expanding the application of deep learning for malware detection, Sweety et al. proposed a multi-view CNN model for both classification and signature generation, pointing to the strength of synergistically integrating behavioral views \cite{Sweety2024}. Divya presented a system named Mal Class, which employs deep CNNs for malware image classification with promising accuracy over various malware families \cite{Divya2025}. Darwish and Roy compared federated learning, traditional machine learning, and deep learning in a study that showed that deep models such as CNNs outranked traditional means under all circumstances, especially with distributed or in IoT-based scenarios \cite{Darwish2025}.

In this study, we propose a CNN-based visual malware classification system that leverages dynamic behavioral data. By capturing runtime API call sequences and transforming them into images, we aim to extract discriminative features for accurate malware classification. The scope of our work includes preprocessing API logs, designing a CNN architecture tailored for malware patterns, and benchmarking the model’s efficiency on real-world data. Our objective is to propose an effective, scalable, and adaptable malware detection framework suitable for deployment in evolving cyber threat landscapes.

The objective of this paper is to explore the effectiveness of using dynamically generated API call-based images to classify malware samples using a CNN architecture. By leveraging the power of visual data and deep learning, we aim at enhancing the robustness and accuracy of malware classification systems. The approach is further validated using a comprehensive experimental setup and evaluation metrics, contributing to the growing domain of visual malware classification research.

The remaining sections of the paper are formatted as follows: Section \ref{sec:basicconcepts} outlines the basic concepts related to Malware Analysis and CNN. Section \ref{sec:relatedwork} reviews the existing related work. Section \ref{sec:framework} details the methodology used in this research. Section \ref{sec:experiment} describes the experimental setup employed. Section \ref{sec:results} presents the results and their analysis.
In Section \ref{sec:compare}, we compare our approach with state-of-the-art techniques. Finally, Section \ref{sec:future} concludes with future research directions.

\section{BASIC CONCEPTS}\label{sec:basicconcepts}

\subsection{Malware Analysis}

Malware analysis is the practice of assessing malware to determine its intent, functional attributes, and potential risk. Security professionals can identify the various strains of malware, come up with appropriate countermeasures, and reinforce the defenses against cyber threats by studying malware carefully. There are two general categories into which the analysis techniques can be classified.

\subsubsection{Static analysis}

Static analysis refers to the process of examining a malware sample without running it. The method analyzes the internal structure of the code, such as embedded strings, file headers, and binary patterns. This aims to distill useful information (e.g., function calls, file dependencies, well-known signatures) that is helpful for enabling early detection.

\subsubsection{Dynamic analysis}

In contrast, dynamic analysis monitors malware behavior when it is operating in a controlled or sandboxed environment. Analysts can use this to watch how the operating system is affected by the malicious code, what it alters, and what external connections it tries to make. Such a real-time glimpse provides valuable insights into the malware’s goals and potential destruction.

\subsection{Sandboxing}

Sandboxing refers to a security method whereby one runs suspected files or software in an isolated environment to identify their behavior without putting the actual system in danger. This approach revolutionizes the detection of zero-day exploits and the analysis of yet unrecognized malware variants. The sandbox emulates a real OS so that malware can behave naturally, but it is also isolated from the main machine so it cannot do any damage.

\subsubsection{Cuckoo Sandbox Analysis Process}

An open-source automated malware analysis tool called Cuckoo Sandbox enables dynamic executable analysis. When a file is executed inside the sandbox, Cuckoo collects information about its behavior and outputs a comprehensive behavioral report in JSON format. It contains information about API calls (invoked by the executable) with their order and frequency, arguments, process-level activity, network traffic, and regime changes. This detail is key for determining the malware's objective, spread methodology, and the damage that could be done to the target computer.

\subsection{Convolutional Neural Networks (CNN)}

This is followed by processing and analyzing visual data, which is called convolutional neural networks (CNN). They learn hierarchical features from images automatically, utilizing completely linked layers, pooling, and convolutions. Mathematically, the convolution operation, which is the basic building block of CNNs, can be described as:

\begin{equation}
\text{output}(i, j) = \sum_m \sum_n \text{input}(i + m, j + n) \cdot \text{kernel}(m, n)
\label{eq:cnn_convolution}
\end{equation}

In this equation:
\begin{itemize}
  \item $\text{output}(i,j)$ is the output at pixel position $(i,j)$.
  \item $\text{input}(i+m,j+n)$ corresponds to the pixel values in the input image at the specific position that is $m$ and $n$ units away respectively from the reference point given by $(i,j)$.
  \item $\text{kernel}(m,n)$ is the filter (or kernel) applied at position $(m,n)$ over the input image.
\end{itemize}

More specifically, a kernel is a matrix of arbitrary size that is used to slide over the image and take dot products as it moves around. CNNs are particularly powerful in fields like image classification and object detection because they can automatically recognize spatial hierarchies, minimizing the need for manual feature extraction for efficient learning from images.

The mapping of features to their images is \textsc{IMAGE}[EXEMPLAR][DENSITY][LIGHT][ANGLE], which indicates that each feature is represented as an image in image space.

\textbf{Experimental Conversion of Features into Image: Data Conversion}

The process included feature encoding to the matrix or image-like structure so that convolutional networks could be applied to the data. A very common function that maps a feature $x$ to another one could be summed up with the following equation:

\begin{equation}
I(x, y) = f(\mathbf{X})
\label{eq:feature_conversion}
\end{equation}

In this equation:
\begin{itemize}
  \item $I(x,y)$: pixel solution on the coordinates $(x,y)$ of the converted image.
  \item $\mathbf{X}$: feature vector derived from the original high-dimensional data.
  \item $f(\mathbf{X})$ represents a function that manipulates the feature vector to give it an image-like appearance (through reshaping or normalizing it into a 2D grid).
\end{itemize}

Because CNNs find underlying spatial patterns in images, features can also be generalized using this data transformation and applied to any type of features which will allow CNN to learn the basic spatial patterns and structure relationships that are characteristic of the features, thus greatly aiding any prediction or analysis task.

\section{RELATED WORK}\label{sec:relatedwork}

Deep learning, particularly with Convolutional Neural Networks (CNNs), has drastically enhanced malware classification, primarily due to their exceptional ability to identify visual trends in the data. Traditional techniques like signature and heuristic-based methods can hardly detect obfuscated and polymorphic malware. Consequently, researchers are now increasingly targeting dynamic, image-based, and hybrid detection methods.

Sasikala and Shanmuganathan~\cite{Sasikala2024} proposed an image-based approach for malware classification based on a specially trained CNN. The approach converts malware binaries into both grayscale and RGB photo format imprints, which gets past the network and truly recognizes design. This method reached 10\% level of accuracy and was resilient to many obfuscation techniques.

Sweety et al.~\cite{Sweety2024} recently proposed a multi-view CNN-based system for dynamic malware detection and signature generation. By analyzing malware behavior from multiple perspectives, their method was able to accurately detect threats and generate precise signatures, highlighting the power of behavior-driven visual analysis.

Divya~\cite{Divya2025} presented \textit{Mal Class}, a CNN-based architecture for automated classification of malware images. Using both monochrome and RGB image representations, the model achieved promising results on the Malimg dataset, showcasing CNNs' capacity for learning discriminative visual patterns even under adversarial variance.

Musaev et al.~\cite{Musaev2025} proposed an ensemble method that integrates a custom CNN with a fine-tuned MobileNetV2. Their Optimized Epoch Selection strategy selects the best-performing model checkpoints, resulting in improved generalization and setting a new accuracy benchmark of 99.05\% on the Malimg dataset.

Wu et al.~\cite{Wu2024} developed MVC-RSN, a malware classification method with variant identification capability. It utilizes ResNet-based architectures to improve generalization and robustness in detecting evolving malware variants, particularly in adversarial settings.

To address the limitations of computational overhead, Yuan et al.~\cite{Yuan2022} designed a lightweight CNN (LCNN) optimized for IoT malware classification. Their approach utilized multidimensional Markov images derived from raw binaries and demonstrated over 99.35\% accuracy on resource-constrained platforms.

Darwish and Roy~\cite{Darwish2025} performed a comparative analysis of federated learning, deep learning, and traditional methods for IoT malware detection. Their findings indicated that deep CNNs consistently outperformed other models in distributed environments, underscoring the value of deep representations.

Javed and Amjad~\cite{Javed2024} explored the integration of dynamic behavioral analysis and SIEM systems with deep learning. Their approach processed API sequences extracted via Cuckoo Sandbox~\cite{CuckooSandbox2025} and achieved significant accuracy improvements through few-shot learning, showcasing the benefit of incorporating contextual behavioral logs.

Gond et al.~\cite{Gond2024} proposed a deep learning framework for malware classification that integrates Natural Language Processing (NLP) techniques. By using n-grams of API call sequences, they successfully capture malware behavior patterns, achieving enhanced classification accuracy and robustness in contrast to conventional techniques.

Rajneekant et al.~\cite{Rajneekant2024} conducted a comparative analysis of machine learning models based on API sequences for malware classification. Their results show that XGBoost outperforms other models, achieving an accuracy of 98.87\%, demonstrating the effectiveness of incorporating API call sequences and arguments for precise malware detection.

Kishore et al.~\cite{Kishore2022} proposed a hybrid analysis model for classifying malicious applications using computationally efficient machine learning techniques. Their method, which combines static and dynamic analysis, achieves high Matthews Correlation Coefficients (MCC), including 89\% to classify malware and 81\% to classify malware families, addressing class imbalance effectively.

Our work expands these core studies by creating a flexible malware classification system that turns API call sequences into grayscale images. We use a deep CNN model trained on these images to effectively pull out features and identify malware families. This blended method enhances both structural and behavioral detection while staying strong against evasion tactics.

\section{PROPOSED FRAMEWORK}\label{sec:framework}
\begin{figure*}[h]
    \centering
    \includegraphics[width=18cm]{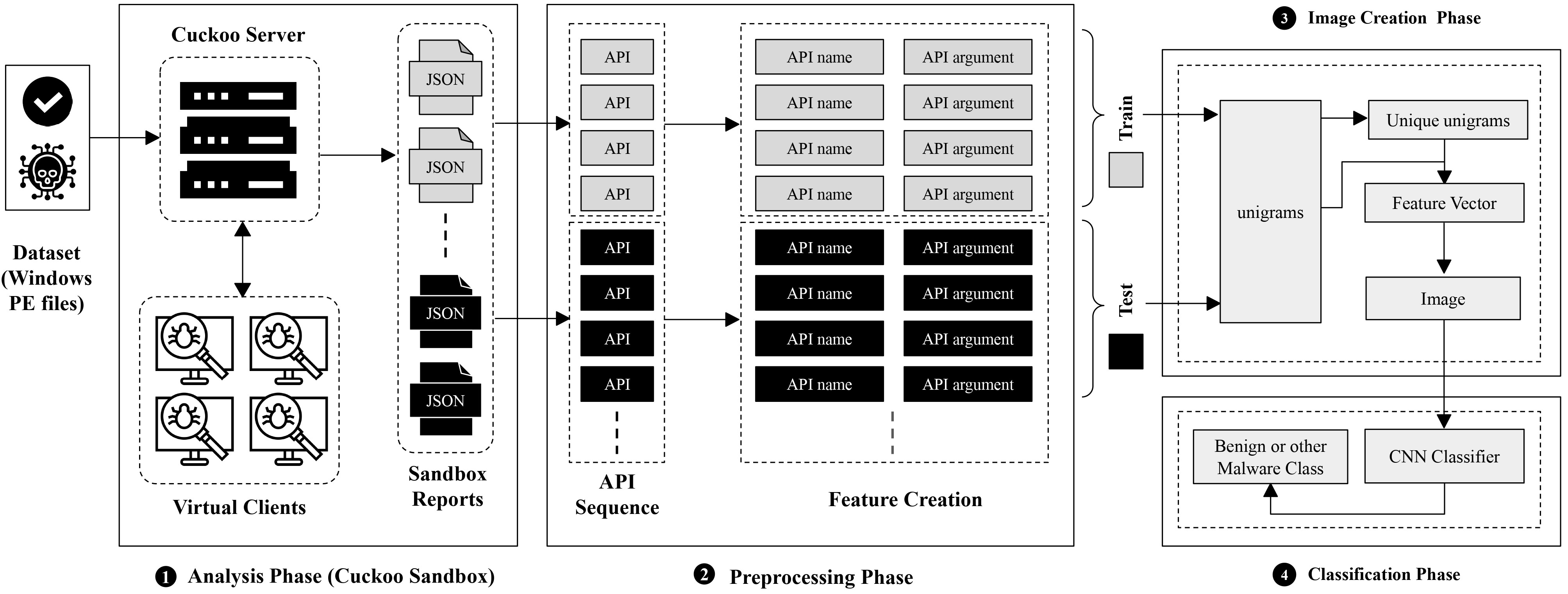}
    \caption{Proposed CNN-based Architecture for Malware Analysis}
    \label{fig:architecture}
\end{figure*}
Figure~\ref{fig:architecture} illustrates our proposed work, which uses a multi-step system to identify malware by analyzing its behavior through image-based deep learning. The process involves preparing malware behavior logs, converting features into visual images, and training a Convolutional Neural Network (CNN) to classify them. The three key stages are: Preprocessing, Feature Transformation, and Model Training \& Evaluation.

\vspace{0.2cm}
\noindent\textbf{Phase 1: Analysis Phase} \\
 In Analysis phase, We carry out the following activities:
 After obtaining the PE files from VirusShare\footnote{\url{https://virusshare.com/}} and Virustotal\footnote{\url{https://www.virustotal.com/}} for all eight multiclass like Adware, Backdoor, Benign, Downloader, Spyware, Trojan, Virus, and Worm we performed behavioral analysis using Cuckoo
 Sandbox, resulting in a behavioral report in JSON format. We then split the file into four parts: API category,
 API name, API argument, and API return. From this,
 We selected the API name and argument to create unigrams, where the API name is the first part and the
 API call argumant is added using underscores.

\textbf{Phase 2: Preprocessing Phase} \\
The following actions are taken during the preprocessing stage:

1) \textbf{API Dataset \& Cuckoo Sandbox Analysis:}  
The data used in this study was collected from VirusShare, a well-known malware sample repository. Each sample was executed in a controlled environment using Cuckoo Sandbox, which generated behavioral reports in JSON format, capturing runtime interactions of PE (Portable Executable) files.

2) \textbf{API Feature Extraction:}  
From each behavioral JSON file, we extracted API-level elements such as \texttt{APICategory}, \texttt{APIName}, \texttt{APIArgument}, and \texttt{APIreturn}. These fields provide insight into how the malware interacts with the system.

3) \textbf{CSV File Construction:}  
For every malware sample, a single row was created in a CSV file. The first column contains the malware hash (unique identifier), the second column specifies the malware family label (e.g., backdoor), and the remaining columns store numeric values representing frequency or intensity of specific API calls.

This CSV file was then used as the structured input for further processing.

\vspace{0.2cm}
\noindent
\textbf{Phase 3: Feature Transformation Phase} \\
This phase focuses on converting the structured CSV data into images\footnote{\href{https://github.com/md-shahnawaz-cse/DMC-MAL-CNN}{Code and Dataset}}
 suitable for training with CNN.

1) \textbf{Normalization and Reshaping:}  
All numeric API values in each row were normalized into a range of 0--255. The resulting vectors were reshaped into square matrices (e.g., 128$\times$128), generating grayscale image representations of API usage patterns for each malware.

2) \textbf{Image Enhancement Techniques:}  
To highlight important features and patterns in the image, the following techniques were applied:
\begin{itemize}
    \item \textbf{Gaussian Blur} -- To reduce noise and smooth variations.
    \item \textbf{CLAHE (Contrast Limited Adaptive Histogram Equalization)} -- For improving local contrast.
    \item \textbf{Sobel Edge Detection} -- To emphasize edge transitions and structural boundaries.
\end{itemize}
\begin{figure}[htbp]
    \centering
    \includegraphics[width=0.9\columnwidth]{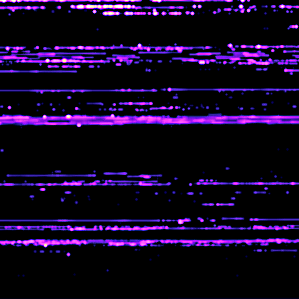}
    \caption{A sample image after applying Magma Colormap}
    \label{fig:sample}
\end{figure}
\begin{figure*}[h]
    \centering
    \includegraphics[width=18cm]{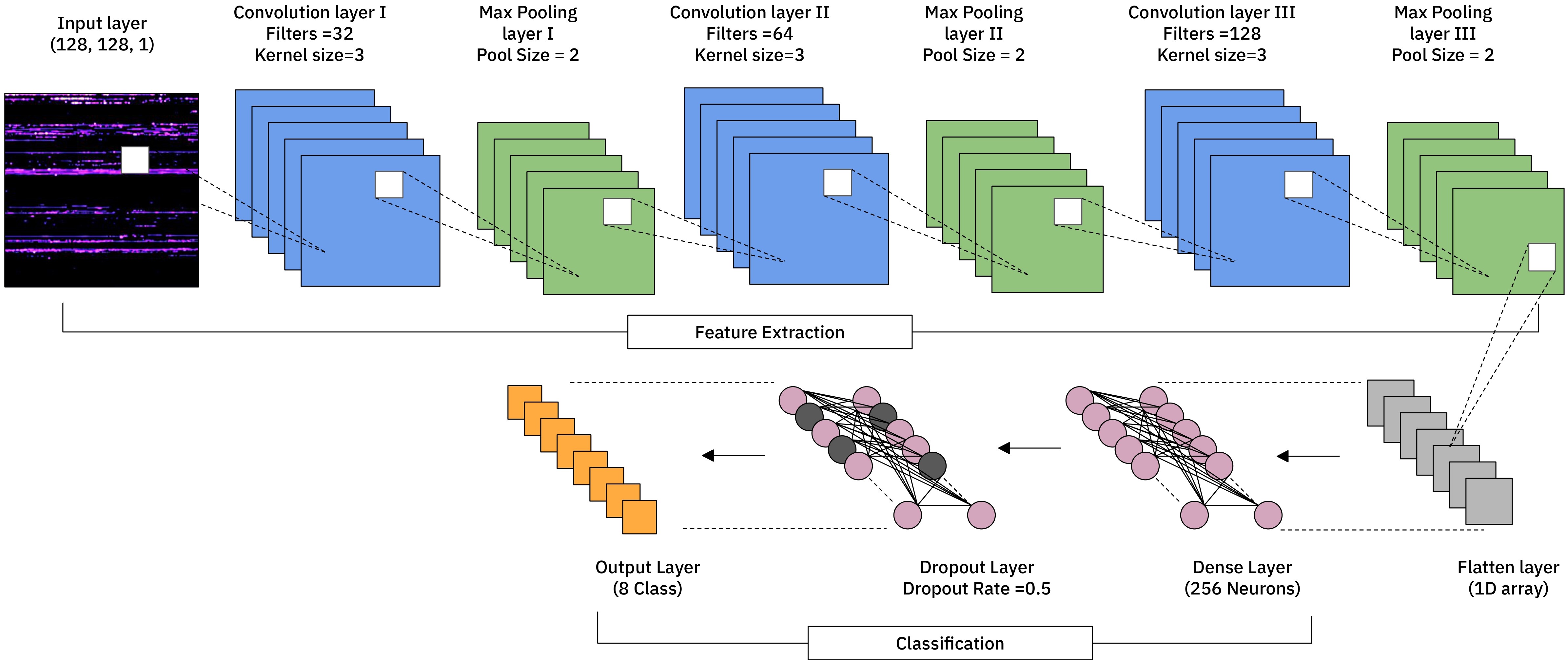}
    \caption{CNN Model Architecture}
    \label{fig:cnnarchitecture}
\end{figure*}
3) \textbf{Color Mapping and Saving:}  
A magma colormap was applied to the grayscale images, adding color richness for enhanced feature representation as shown in Figure~\ref{fig:sample}. Finally, contrast and sharpness improvements were applied before saving the images to disk for training.


\textbf{Phase 4: Model Training and Evaluation Phase} \\

This phase includes the design, training, and evaluation of a CNN model tailored for malware classification using grayscale images generated from API call patterns.

\subsubsection{\textbf{CNN Model Architecture}}
The architecture for our CNN model is defined as follows:
\begin{align*}
    \textbf{Input on layer:} &\ (128, 128, 1) \\
    \textbf{Convolutional Layer 1:} &\ Filters = 32,\ \text{Kernel Size} = 3,\\ &\text{Activation} = \text{ReLU} \\
    \textbf{MaxPooling Layer 1:} &\ \text{Pool Size} = 2 \\
    \textbf{Convolutional Layer 2:} &\ Filters = 64,\ \text{Kernel Size} = 3,\\ &\text{Activation} = \text{ReLU} \\
    \textbf{MaxPooling Layer 2:} &\ \text{Pool Size} = 2 \\
    \textbf{Convolutional Layer 3:} &\ Filters = 128,\ \text{Kernel Size} = 3,\\ & \text{Activation} = \text{ReLU} \\
    \textbf{MaxPooling Layer 3:} &\ \text{Pool Size} = 2 \\
    \textbf{Flatten Layer:} &\ \text{Flattens the input to a 1D array} \\
    \textbf{Dense Layer:} &\ \text{Neurons} = 256,\ \text{Activation} = \text{ReLU} \\
    \textbf{Dropout Layer:} &\ \text{Dropout Rate} = 0.5 \\
    \textbf{Output Layer:} &\ \text{Neurons} = 8,\ \text{Activation} = \text{Softmax}
\end{align*}

\textbf{CNN Architecture:}  
We designed a custom CNN from scratch as shown in Figure~\ref{fig:cnnarchitecture}, comprising:
\begin{itemize}
    \item Three convolutional layers with increasing filter sizes.
    \item Three Max pooling layers for downsampling after each convolution.
    \item A flattened layer followed by a dense layer with dropout for regularization.
\end{itemize}

The images were resized to 128$\times$128 pixels and normalized before feeding into the network. To improve generalization, during training, data augmentation methods like flipping, zooming, and rotation were used.

\subsubsection{\textbf{Training and Output Generation:}  }
The CNN model was trained for 100 epochs utilizing categorical cross-entropy loss and the Adam optimizer. The model artifacts, including trained weights, prediction results, evaluation plots, and training logs, were saved in a time-stamped results folder.

\subsubsection{\textbf{Performance Evaluation:}  }
The trained model got evaluated using
\begin{itemize}
    \item \textbf{Confusion Matrix:} Analyzing False Negatives, True Negatives, True Positives, and False Positives.
    \item \textbf{Classification Metrics:} Including F1-Score, Accuracy, Precision, and Recall.
\end{itemize}

These metrics collectively provided thorough explanation of how well the model could categorize different malware families.


\section{EXPERIMENTAL SETUP} \label{sec:experiment}

Our experimental setup was designed to evaluate the effectiveness of image-based malware classification using behavioral API features. It comprises the following components:

1) \textbf{Analysis Environment:}  
\begin{itemize}
    \item \textbf{Host System:} An Intel Xeon(R) Silver 4216 CPU with 128 GB RAM and 5TB HDD was used for executing the Cuckoo Sandbox, performing preprocessing, and training the deep learning model.
\end{itemize}

2) \textbf{Windows 10 Environment:}  
\begin{itemize}
    \item \textbf{OS:} Windows 10  
    \item \textbf{Hardware:} A machine with a 5TB storage capacity, 128GB of RAM, an Intel i7 processor, and a 16GB NVIDIA GPU was utilized to collect and analyze the dynamic behavior of Cuckoo Sandbox reports.
\end{itemize}

3) \textbf{Development Environment:}  
\begin{itemize}
    \item \textbf{Programming Language:} Python Python 3.10.9 was used to implement the complete pipeline from data extraction to model evaluation.
    \item \textbf{IDE:} Spyder IDE facilitated efficient development, debugging, and visualization.
\end{itemize}

\begin{table}[ht!]
\caption{Datasets used}
\begin{center}
\begin{tabular}{|c|c|c|c|c|}
\hline
\textbf{S.No} & \textbf{Types} & \textbf{Test Sample} & \textbf{Train Sample} & \textbf{Total Sample} \\
\hline\hline
1 & Adware & 316 & 1670 & \textbf{1986} \\
\hline
2 & Backdoor & 228 & 446 & \textbf{674} \\
\hline
3 & Downloader & 500 & 1999 & \textbf{2499} \\
\hline
4 & Spyware & 167 & 779 & \textbf{946} \\
\hline
5 & Trojan & 675 & 2893 & \textbf{3568} \\
\hline
6 & Virus & 464 & 1928 & \textbf{2392} \\
\hline
7 & Worms & 305 & 1052 & \textbf{1357} \\
\hline
8 & Benign & 1857 & 6777 & \textbf{8634} \\
\hline\hline
 & \multicolumn{1}{|c|}{\textbf{Total}} & \textbf{4512} & \textbf{17544} & \textbf{22056} \\
\cline{2-5} 
\hline
\end{tabular}
\label{tab:mal_data}
\end{center}
\end{table}

4) \textbf{Malware Samples:}  
\begin{itemize}
    \item A dataset consisting of 22,056 samples \cite{gond2025malware} was used, out of which 17,544 samples were utilized for training and 4,512  samples for testing the classification model as shown in Table \ref{tab:mal_data}.
\end{itemize}

This setup facilitated controlled experimentation to evaluate the potential of converting API sequences into visual features for improving malware classification performance. All steps—from Cuckoo-based behavior capture to CNN-based classification—were performed within this environment.

\section{RESULT ANALYSIS}\label{sec:results}

In this section, we evaluate the performance of the proposed malware classification model using a confusion matrix, detailed performance metrics, and a visual comparison of classification outcomes across various malware types.

\subsection{Confusion Matrix Analysis}
\begin{figure}[htbp]
    \centering
    \includegraphics[width=9cm]{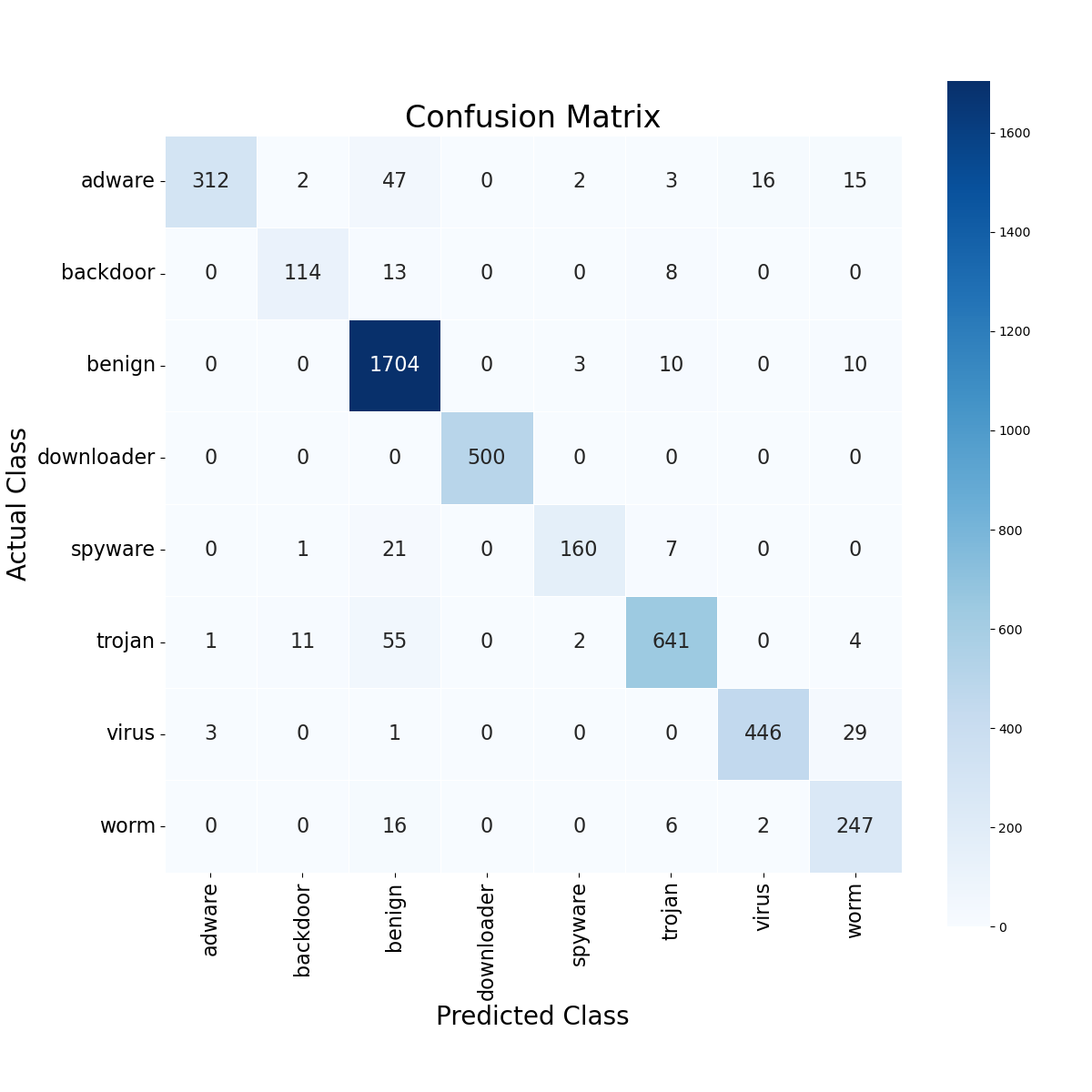}
    \caption{Confusion Matrix of malware classification}
    \label{fig:confusion3}
\end{figure}

Figure~\ref{fig:confusion3} illustrates the confusion matrix, providing insight into the model’s performance across eight malware types and benign samples. The model performs exceptionally well for the \textit{benign} and \textit{downloader} classes, achieving 1704 and 500 true positives, respectively, with almost negligible misclassifications. For the \textit{trojan} class, the model correctly predicts 641 samples, but there are a few misclassifications into \textit{benign} and \textit{backdoor}. Classes like \textit{adware} and \textit{worm} show moderate confusion with other categories, highlighting areas where the model can benefit from further optimization or additional feature refinement.

\subsection{Performance Metrics Overview}
\begin{table}[htbp]
    \centering
    \caption{Performance metrics for each malware type}
    \label{tab:performance_metrics_malware}
    \begin{tabular}{|c|c|c|c|c|c|}
        \hline
        \textbf{No} & \textbf{Malware Type} & \textbf{Accuracy} & \textbf{F1 Score} & \textbf{Recall} & \textbf{Precision} \\
        \hline \hline
        1 & Adware     & 98.00\% & 87.50\% & 78.60\% & 98.70\% \\ \hline
        2 & Backdoor   & 99.20\% & 86.70\% & 84.40\% & 89.10\% \\ \hline
        3 & Benign     & 96.00\% & 95.10\% & 98.70\% & 91.80\% \\ \hline
        4 & Downloader & 100.00\% & 100.00\% & 100.00\% & 100.00\% \\ \hline
        5 & Spyware    & 99.20\% & 89.90\% & 84.70\% & 95.80\% \\ \hline
        6 & Trojan     & 97.60\% & 92.30\% & 89.80\% & 95.00\% \\ \hline
        7 & Virus      & 98.80\% & 94.60\% & 93.10\% & 96.10\% \\ \hline
        8 & Worm       & 98.10\% & 85.80\% & 91.10\% & 81.00\% \\
        \hline
    \end{tabular}
\end{table}
\begin{figure}[htbp]
    \centering
    \includegraphics[width=9cm]{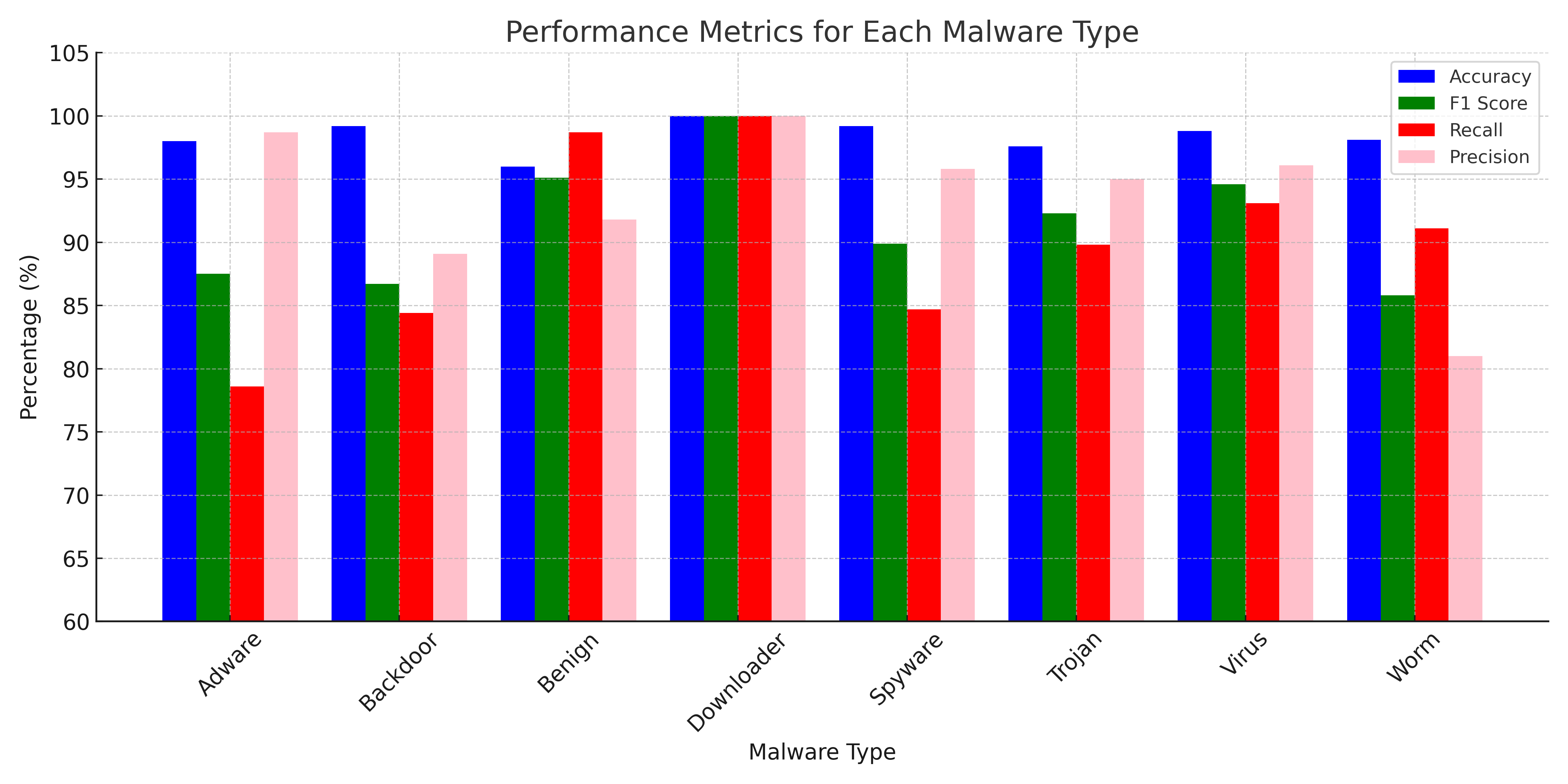}
    \caption{Bar graph comparing Accuracy, F1 Score, Recall, and Precision for each malware type}
    \label{fig:bar_metrics}
\end{figure}

As shown in Table~\ref{tab:performance_metrics_malware}, the model performs with high accuracy across all categories, with \textit{downloader} achieving a perfect 100\% in all metrics. \textit{Benign} and \textit{virus} categories also exhibit outstanding F1 scores and recall, indicating reliable detection with minimal false negatives.

Figure~\ref{fig:bar_metrics} provides a visual summary of performance, confirming the consistent and balanced results. It is evident that while \textit{adware}, \textit{spyware}, and \textit{backdoor} show slightly lower recall, their precision remains high, suggesting correct positive detections when identified.

\begin{figure}[htbp]
    \centering
    \includegraphics[width=9cm]{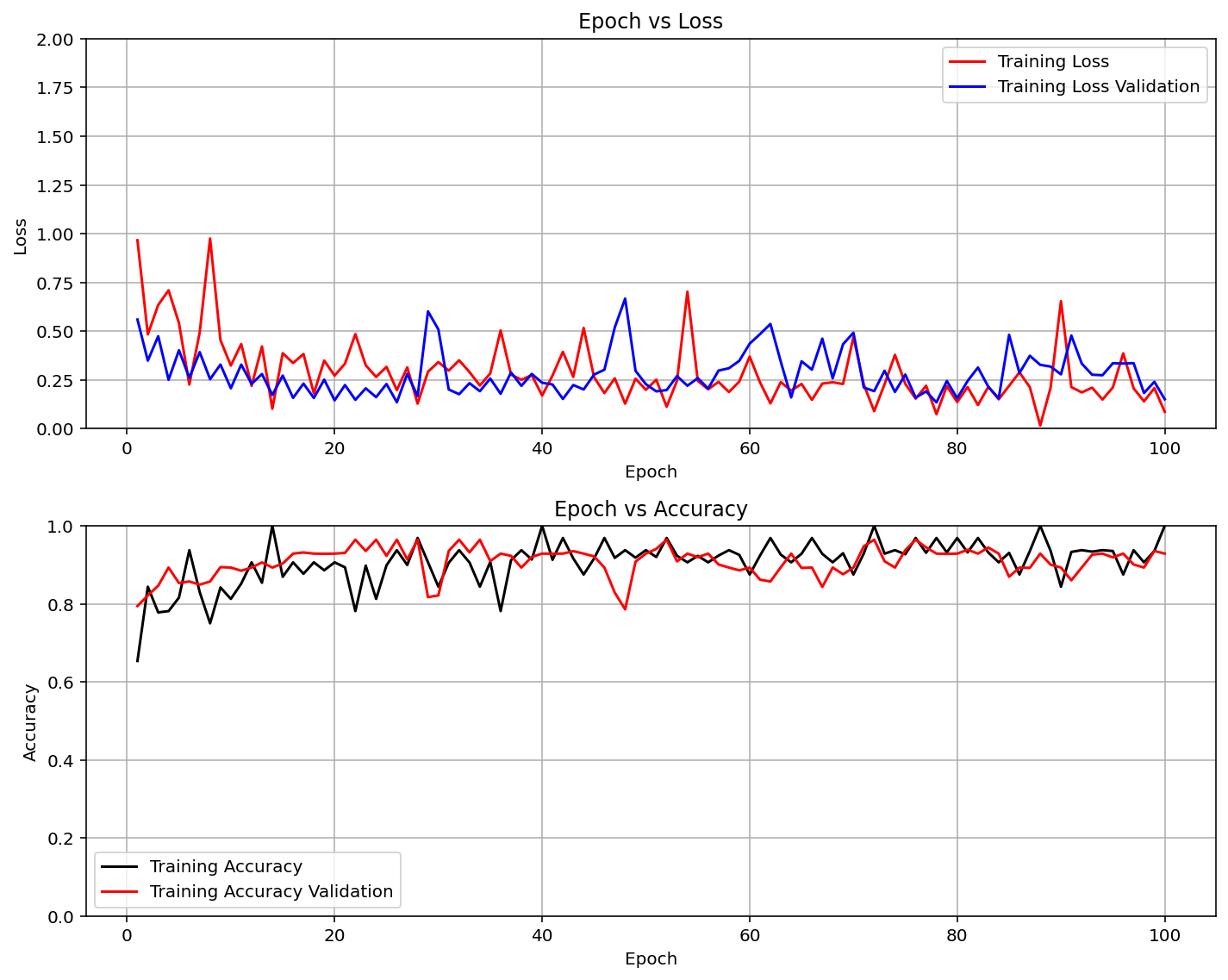}
    \caption{Loss and Accuracy measures for proposed model}
    \label{fig:lossacc}
\end{figure}

\begin{table}[htbp]
    \centering
    \caption{Evaluation metrics of malware classification}
    \label{tab:tp_tn_fp_fn_malware}
    \begin{tabular}{|c|c|c|c|c|c|}
        \hline
        \textbf{S. No.} & \textbf{Malware Type} & \textbf{TP} & \textbf{TN} & \textbf{FP} & \textbf{FN} \\
        \hline \hline
        1 & Adware     & 312  & 4011 & 4   & 85 \\ \hline
        2 & Backdoor   & 114  & 4263 & 14  & 21 \\ \hline
        3 & Benign     & 1704 & 2532 & 153 & 23 \\ \hline
        4 & Downloader & 500  & 3912 & 0   & 0  \\ \hline
        5 & Spyware    & 160  & 4216 & 7   & 29 \\ \hline
        6 & Trojan     & 641  & 3664 & 34  & 73 \\ \hline
        7 & Virus      & 446  & 3915 & 18  & 33 \\ \hline
        8 & Worm       & 247  & 4083 & 58  & 24 \\
        \hline
    \end{tabular}
\end{table}

Figure~\ref{fig:lossacc} shows, our model exhibits effective learning dynamics. The training and validation loss steadily decrease over epochs, indicating proper convergence. The accuracy plot reveals a high training accuracy and a consistently stable validation accuracy, suggesting that the model generalizes well. Minimal oscillations in validation accuracy imply controlled overfitting and good performance across unseen data.

\subsection{True/False Positive-Negative Metrics}

Table~\ref{tab:tp_tn_fp_fn_malware} summarizes the actual classification counts. Again, \textit{downloader} is the top-performing class with perfect TP and TN and no FP or FN. Conversely, the \textit{benign} class shows the highest false positive count (153), which may indicate that some benign samples resemble malware patterns.

In summary, the proposed CNN-based approach demonstrates strong classification performance, especially for key classes like \textit{virus}, \textit{downloader}, and \textit{benign}. While minor recall dips are noted for a few malware types, the overall results confirm the model's effectiveness for real-world malware classification.


\section{COMPARISON OF OUR WORK WITH PRESENT STATE-OF-THE-ART TECHNIQUES}\label{sec:compare}

Our proposed approach focuses on classifying malware through dynamic analysis by converting API call sequences into grayscale images and leveraging a Convolutional Neural Network (CNN) for classification. While our dataset and image-generation method are unique, and hence, direct comparison is limited, we present a comparative discussion with existing deep learning-based techniques to highlight the effectiveness and distinctiveness of our method in Table~\ref{tab:comparison}.

Sweety et al.~\cite{Sweety2024} developed a multi-view CNN framework that analyzes malware from several behavioral perspectives. Their approach effectively classified and generated malware signatures, leveraging dynamic execution features collected from sandbox environments. Although comprehensive in behavior analysis, their architecture is relatively more complex compared to our single-view grayscale image approach.

Divya~\cite{Divya2025} introduced \textit{Mal Class}, a CNN-based deep learning method for classifying malware images. Their system used RGB and monochrome images converted from malware binaries. While the approach is similar in using image inputs, our work differs by utilizing behavioral API sequences rather than static binaries, thereby improving robustness against polymorphism.

Musaev et al.~\cite{Musaev2025} proposed an ensemble of custom CNN and fine-tuned MobileNetV2 for image-based malware classification. Their technique, optimized with epoch selection, achieved 99.05\% accuracy on the Malimg dataset. Our method, while simpler in architecture, is tested on a much larger and dynamically generated dataset, showcasing practical adaptability.

Wu et al.~\cite{Wu2024} developed MVC-RSN, a ResNet-based malware classification model with variant identification capabilities. Their work excels in generalization across adversarial malware, whereas our work emphasizes behavioral pattern capture through visual encoding of API sequences.

Yuan et al.~\cite{Yuan2022} focused on lightweight CNNs for IoT malware detection. Their method achieved high accuracy with minimal resource usage, targeting constrained devices. Our method prioritizes robustness and behavioral coverage, suitable for enterprise-level malware detection systems.

\begin{table*}[hbt!]
\caption{Quantitative comparison of selected deep learning-based malware detection and classification techniques}
\begin{center}
\begin{tabular}{ |c|c|c|c|c|c|c|c|l| }
 \hline
 \textbf{No} &\textbf{Author} &\textbf{Image Type} &\textbf{Deep Learning Model} &\textbf{Dataset Used} &\textbf{Dataset Size} &\textbf{Detection} &\textbf{Cls\textsuperscript{n}
} &\textbf{Acc} \\
 \hline
 \hline
1 & Musaev et al.~\cite{Musaev2025} & Grayscale & Custom CNN + MobileNetV2 & Malimg~\cite{MalimgDataset}  & 9339 & \ding{55} & \ding{51} & 99.05\% \\ \hline
2 & Wu et al.~\cite{Wu2024} &  Binary  & MVC-RSN (ResNet-based) & Custom & Not disclosed & \ding{55} & \ding{51} & 98.6\% \\ \hline
3 & Yuan et al.~\cite{Yuan2022} &  Markov Images & Lightweight CNN & VirusShare\cite{VirusShare2025}  & Not disclosed & \ding{51} & \ding{55} & 99.36\% \\ \hline
4 & Sweety et al.~\cite{Sweety2024} & Feature Maps & Multi-View CNN & Custom & 6000+ & \ding{51} & \ding{51} & 98\% \\ \hline
5 & Divya~\cite{Divya2025} & Grayscale / RGB & CNN & Malimg~\cite{MalimgDataset} & Not disclosed & \ding{55} & \ding{51} & 98.99\% \\ \hline
6 & Darwish and Roy~\cite{Darwish2025} & Not Image-Based & CNN / Federated Learning & NetworkX & 6500+ & \ding{51} & \ding{51} & 95.27\% \\ \hline
7 & Proposed Work & Greyscale & CNN & VirusShare\cite{VirusShare2025} & \textbf{22056} & \ding{55} & \ding{51} & 98.36\% \\
\hline
\end{tabular}
\label{tab:comparison}
\end{center}
\end{table*}

Darwish and Roy~\cite{Darwish2025} presented a comparative evaluation of federated learning, deep learning, and traditional models for IoT malware detection. Their findings reinforce the strength of CNNs in distributed environments, supporting the foundational design of our image-based CNN model.

Our methodology stands apart due to its combination of runtime API call sequence analysis, $n$-gram extraction, grayscale image conversion, and CNN-based classification. We evaluated our model on a dataset of 22,056 samples spanning seven malware categories, capturing behavioral semantics through visual data. This hybrid approach makes our system robust, scalable, and suitable for deployment in dynamic cybersecurity environments.

\section{CONCLUSION AND FUTURE WORKS}\label{sec:future}

In this research, we proposed a new malware classification approach that transforms dynamically captured API call sequences into grayscale images and classifies them using a Convolutional Neural Network (CNN). By leveraging behavioral data and visual pattern recognition, our model effectively identified and classified seven major malware families across a dataset comprising 22,056 samples. The results demonstrated high classification performance, particularly for malware types such as Downloader, Backdoor, Spyware, Worms, Adware, Trojan, and Virus, with accuracy scores exceeding 97\%. This image-based behavioral modeling provides a scalable and effective alternative to traditional static analysis methods, especially in handling obfuscation and evasive malware strategies.

Our study demonstrates the potential of integrating dynamic execution data with deep learning for robust malware detection. By converting API sequence patterns into visual representations, the model captures both structural and temporal features, enabling it to generalize across varied malware classes.

\vspace{0.5em}
\noindent Below are some future directions to enhance this research:

\begin{itemize}
    \item \textbf{LLM-Assisted API Interpretation:} Integrating Large Language Models (LLMs) to semantically analyze API arguments and contextual behaviors can enhance feature richness and interpretability, especially for novel or evolving threats.

    \item \textbf{Cross-Platform Compatibility:} Extending this framework to classify Android and Linux-based malware using system call equivalents could broaden its applicability across platforms.

    \item \textbf{Model Compression:} Exploring lightweight CNNs and pruning techniques will be important for deploying the model in resource-constrained environments like IoT gateways or mobile endpoints.

    \item \textbf{Advanced Similarity Learning:} Incorporating Siamese networks or triplet loss could provide a deeper understanding of intra-class and inter-class similarity in malware behaviors.

    \item \textbf{Temporal Sequencing:} Combining CNN with temporal models like LSTM or Transformer-based architectures may improve understanding of long-range behavioral dependencies in complex malware routines.
\end{itemize}

Through these future extensions, we aim to evolve our malware classification system into a highly accurate, adaptive, and lightweight solution capable of defending against next-generation cyber threats in real-time.

\bibliographystyle{unsrt}

\end{document}